\documentclass[letterpaper, 10 pt, conference]{ieeeconf}  

\IEEEoverridecommandlockouts                              

\usepackage{graphics} 
\usepackage{amsmath} 
\usepackage{amssymb}  

\usepackage{graphicx}

\usepackage[x11names, rgb]{xcolor}

\usepackage{subfig}
\usepackage{siunitx}

\title{\LARGE \bf
Efficient End-to-End Detection of 6-DoF Grasps for Robotic Bin Picking
}

\author{Yushi Liu$^{1,2}$ \quad Alexander Qualmann$^{1}$ \quad Zehao Yu$^{2}$ \quad Miroslav Gabriel$^{1}$  
\\Philipp Schillinger$^{1}$ \quad Markus Spies$^{1}$ \quad
Ngo Anh Vien$^{1}$ \quad Andreas Geiger$^{2}$
\thanks{$^{1}$Bosch Center for Artificial Intelligence, Renningen, Germany.}
\thanks{$^{2}$University of Tuebingen, Tuebingen AI Center, Germany.}
}

\begin{document}

\maketitle
\thispagestyle{empty}
\pagestyle{empty}

\begin{abstract}
    Bin picking is an important building block for many robotic systems, in logistics, production or in household use-cases. 
    In recent years, machine learning methods for the prediction of 6-DoF grasps on diverse and unknown objects have shown promising progress.
    However, existing approaches only consider a single ground truth grasp orientation at a grasp location during training and therefore can only predict limited grasp orientations which leads to a reduced number of feasible grasps in bin picking with restricted reachability.
    In this paper, we propose a novel approach for learning dense and diverse \mbox{6-DoF} grasps for parallel-jaw grippers in robotic bin picking. 
    We introduce a parameterized grasp distribution model based on Power-Spherical distributions that enables a training based on all possible ground truth samples. 
    Thereby, we also consider the grasp uncertainty enhancing the model’s robustness to noisy inputs. 
    As a result, given a single top-down view depth image, our model can generate diverse grasps with multiple collision-free grasp orientations. 
    Experimental evaluations in simulation and on a real robotic bin picking setup 
    demonstrate the model’s ability to generalize across various object categories achieving an object clearing rate of around 90\% in simulation and real-world experiments.
    We also outperform state of the art approaches.
    Moreover, the proposed approach exhibits its usability in real robot experiments without any refinement steps, even when only trained on a synthetic dataset, due to the probabilistic grasp distribution modeling.

\end{abstract}

\section{INTRODUCTION}
Grasp detection is an essential problem for the automation of pick-and-place tasks in industry and logistics, where robots are used to grasp and manipulate objects in unstructured environments.
\begin{figure}[t]
        \centering
        \vspace{0.5cm}
        \includegraphics[width=0.99\linewidth]{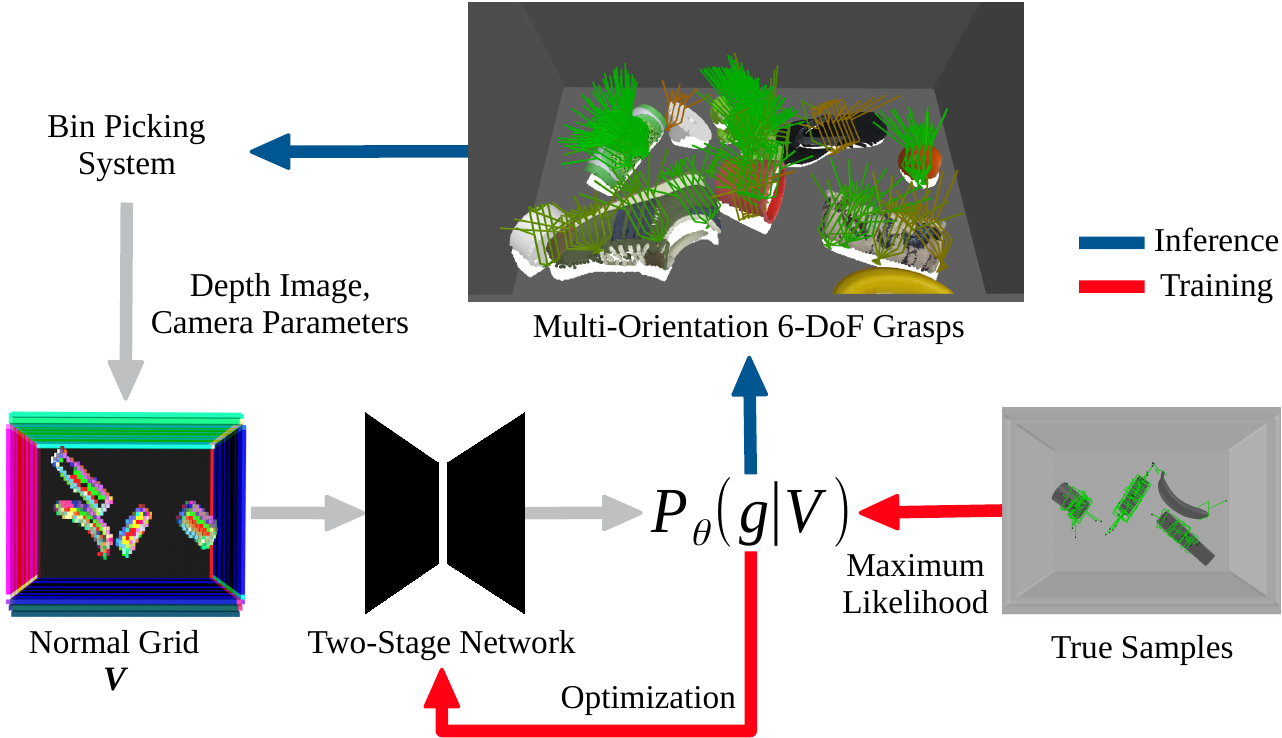}
        \caption{Schematic overview on training and inference pipelines of the proposed \mbox{6-DoF} grasp prediction model. The network is trained to predict the optimal parameter $\theta$ of the grasp distribution $P_\theta(g | V)$, maximizing the likelihood of ground truth grasp samples; During inference, the model infers grasps with multiple orientations including graspability and collision scores. Green forks indicate 6-DoF grasps.}
        \label{fig:detection_model}
\end{figure}
Recent research applies machine learning methods to enable model-free grasp detection for parallel-jaw grippers on diverse and previously unknown objects.
Some approaches address the 4-DoF grasp problem \cite{dexnet2, morrison2018closing}, primarily considering 3D position and gripper orientation about a gravity-aligned approach vector.
While this simplifies the grasping problem, it requires assumptions on the gripper orientation, leading to a top-down grasp execution limitation.
Other approaches propose end-to-end 6-DoF grasp detection learning based on a deep neural network architecture for feature extraction from 3D input data~\cite{sundermeyer2021contactgrasp,qi2017pointnetplusplus, breyer2020volumetric, jiang2021synergies, cai2022vpn}. Even though these approaches directly output multiple complete gripper poses, they only predict one gripper orientation at a location an thus neglect other potential gripper orientations. 
Jeng. et. al. \cite{jeng2021gdn} propose an end-to-end network predicting confidence scores for various grasp orientations, along with refinement values for diverse gripper orientations per grasp point. However, this approach requires a specific fine-tuning process of grasp poses.
Furthermore, current methods only consider a single ground truth grasp orientation at a grasp point during training which then limits the number of feasible grasps during inference and can reduce the grasp performance in bin picking with additional reachability restrictions. 
In this paper, we propose a novel approach for predicting dense parallel-jaw grasps in cluttered bin picking scenarios from single-view observations. Our method predicts multiple orientations per contact point based on grasp distribution modeling.

Our contributions are as follows: 
(1) A novel objective function based on a distribution modeling and integrating uncertainty learning for different grasp orientations at each contact point. The function enables learning with multiple diverse true grasp samples per contact point and enhances the method's robustness to noisy input.
(2) A two-stage end-to-end trainable network for 6-DoF grasp prediction, which generates dense and collision-free grasps for a single-view depth image including multiple grasp orientations at each contact point.
(3) Experimental evaluation in simulation and on a real robot bin picking setup (see Fig.~\ref{fig:real_setup_objects}).
The results show that our method outperforms state of the art baseline methods and demonstrates the adaptability of the approach to real robot bin picking setups, even when only trained on a synthetic dataset, due to the integrated probabilistic grasp distribution modeling.

\section{Related Work}

Grasp detection is a widely researched topic, with notable surveys like~\cite{tian2023datadriven} and~\cite{kleeberger2020survey}. In this section, we focus on the parts of this research most relevant to our work.

\subsection{Data-driven 6-DoF Grasp Detection}

Recent 6-DoF grasp detection techniques directly regress grasp poses using scene sensor data. Many methods treat each surface point as a grasp center and predict grasp parameters for every voxel center~\cite{breyer2020volumetric, jiang2021synergies}, point~\cite{sundermeyer2021contactgrasp, qi2017pointnetplusplus}, or pixel~\cite{gou2021rgbd_grasp}. This approach can be inefficient when the input space significantly exceeds the grasp translation space. Another common strategy~\cite{fang2020graspnet, wang2021graspness, zhao2021regnet, wei2021gpr} uses multi-stage networks to simplify the 6-DoF grasp representation, with each stage predicting specific grasp parameters. These methods typically consider only the closest ground truth grasp during training, limiting the range of predicted orientations. Our method, in contrast, first identifies contact point candidates and then predicts grasp parameters based on these points. Additionally, we define a local grasp orientation distribution for each contact point. This allows our model to train on various grasp labels, producing multiple grasp orientations for each contact. While the work by Jeng et al.\cite{jeng2021gdn} is similar in that it also generates multiple orientations per grasp center using a coarse-to-fine approach, our method stands out by directly outputting accurate grasp poses for each contact point using local grasp distribution modeling, eliminating the need for refinement.

\subsection{Scene Representation}

Scene representation is crucial for efficient grasp prediction. The Truncated Signed Distance Function (TSDF) is a popular choice due to its ability to represent near-surface information. However, generating a comprehensive TSDF grid requires multi-view depth or RGB inputs~\cite{breyer2020volumetric, cai2022vpn,Dai2023GraspNeRF}, which is challenging in bin picking contexts. In contrast, TSDF grids from monocular view often miss essential geometric details, e.g. the contact surface.
While auxiliary tasks such as shape completion or 3D reconstruction can improve predictions~\cite{jiang2021synergies}, we prioritize reconstructing spaces crucial for grasping, including the contact region, to boost training efficiency. As pointed out in~\cite{ten2017grasp, riedlinger2020model, cai2022volumetricbased, eppner2021acronym} that surface normals is of great importance for predicting grasp orientation, our model utilizes a normal grid where each voxel encoding a surface normal.

\subsection{Grasp Pose Representation}

Recent research favors the 6-DoF grasp representation due to its ability to offer greater grasping versatility. These methods primarily utilize points from the point cloud or voxel centers as contact candidates for the gripper's end-point position. \cite{6dof_graspnet} directly maps grasp poses in SE(3), employing grasp translation and a unit quaternion for rotation. Other methods~\cite{cai2022volumetricbased,breyer2020volumetric,jiang2021synergies,liang2019pointnetgpd} adopt a similar representation to predict point-wise grasp configurations. Additionally, methods like~\cite{sundermeyer2021contactgrasp,fang2020graspnet,fang2023anygrasp} use Gram-Schmidt orthonormalization to define grasp poses with an approach and rotation vector. GDN~\cite{jeng2021gdn} proposes learning multiple approach vectors for each feature volume. However, these methods predominantly represent only one grasp per contact point. In contrast, our approach constructs a parameterized grasp orientation distribution at each grasp contact point using a Power-Spherical distribution, enabling infinite ways to grasp objects at a single contact point.


\section{Problem Statement}
We consider the problem of 6-DoF grasp detection for random bin picking scenes with cluttered and unknown objects located inside a bin and captured by a single \mbox{top-down} view camera depth image.


\begin{figure}[!t]
        \centering
        \vspace{0.3cm}
        \includegraphics[width=0.6\columnwidth]{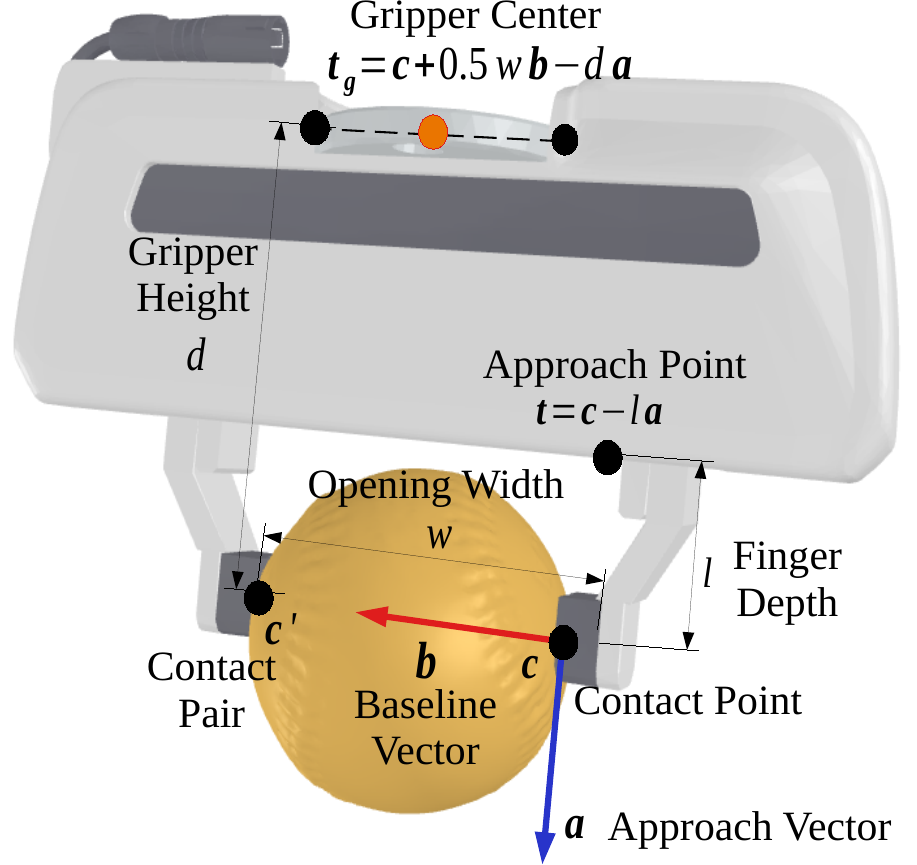}
        \caption{Contact grasp. }
        \label{fig:contact_grasp}
\end{figure}
\subsection{Notations}\label{sec:notations}
\noindent \textbf{3D Normal Grid:}
We define a normal grid, denoted as $\mathbf{V}$, as an $N^3$ voxel grid where each voxel that is close to the object surface stores the normal vector corresponding to the nearest surface point relative to the voxel's center.
To create this 3D normal grid, we initially transform the raw depth image into a truncated signed distance field (TSDF). Subsequently, we extract surface normals for voxel centers located closely to the object surface. These extraction techniques are based on methods described in prior works such as \cite{curless1996volumetric, newcombe2011kinectfusion, Zhou2018}.
\noindent \textbf{Contact Grasp: }We use the contact grasp representation, i.e. the Gram-Schmidt
orthonormalization, for parallel-jaw grippers proposed in \cite{sundermeyer2021contactgrasp}. 
Each contact grasp $g$ at a contact point $\mathbf{c} \in \mathbb{R}^3$ is represented as a 4-tuple: 
\begin{equation}\label{eq:contact_grasp}
 g = (\mathbf{c}, \mathbf{b}, \mathbf{a}, w),
\end{equation} 
where $\mathbf{b} \in \mathbb{R}^3$ is the grasp baseline vector, $\mathbf{a} \in \mathbb{R}^3$ is the grasp approach vector, and $w \in \mathbb{R^+}$ is the grasp opening width of the parallel-jaw gripper (Fig. \ref{fig:contact_grasp}).
        
\noindent \textbf{Approach Point: }A reference point $\mathbf{t}$ of the gripper hand location (see Fig.~\ref{fig:contact_grasp}). We select this point because it is close to the contact point and related to approach vector. We also observed that learning its distribution is beneficial for predicting collision-free approach directions.

\noindent \textbf{Antipodal Quality: }A metric $q$ measuring grasp stability by calculating the \textit{antipodality}~\cite{dexnet2, cai2022volumetricbased, newbury2022review} of the contact pair $\mathbf{(c_1, c_2)}$ of a parallel gripper, which is defined as 
\begin{equation}\label{eq: ap_score}
    q(\mathbf{c_1},\mathbf{c_2}) = |\cos (\mathbf{b}, \mathbf{n(c_1})| |\cos(\mathbf{b}, \mathbf{n(c_2})|,
\end{equation}
where $\mathbf{b}$ is the baseline vector, $\mathbf{n(\cdot)}$ computes the surface normal vectors at two contact points. 

\begin{figure}[!t]
    \centering
    \vspace{0.3cm}
    \includegraphics[width=0.75\columnwidth]{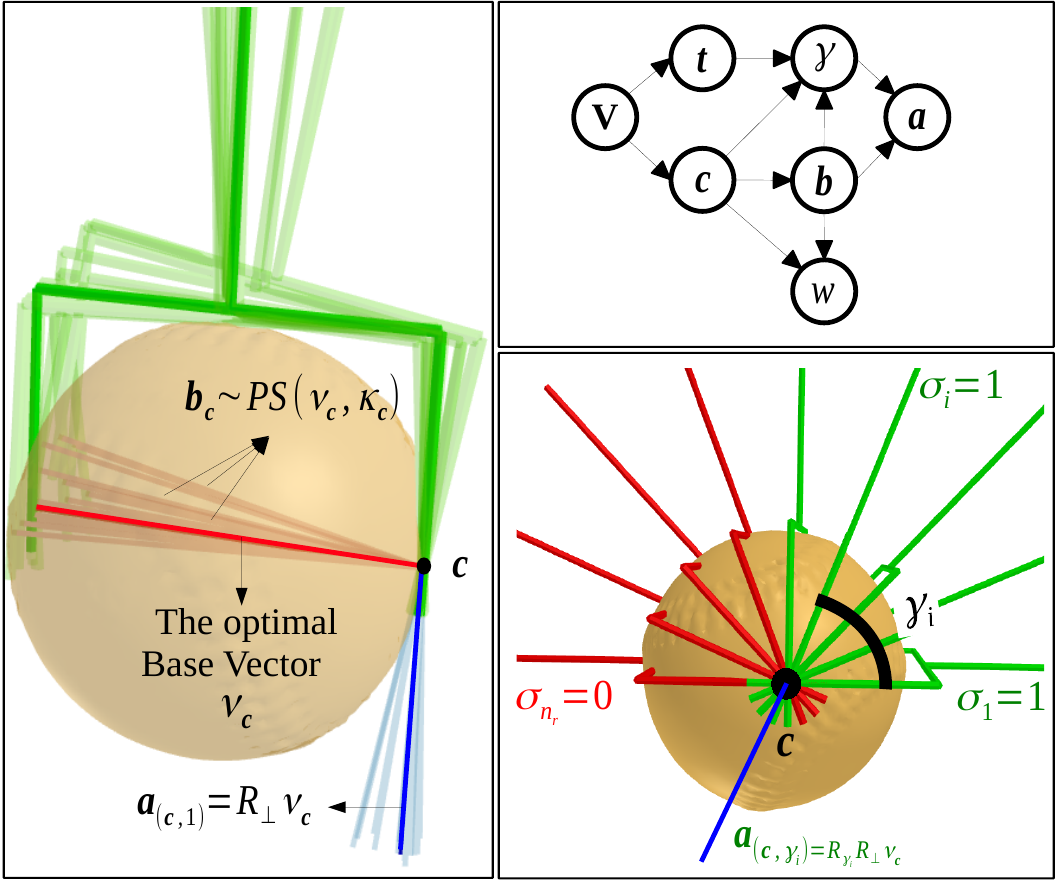}
    \caption{
    A baseline vector distribution (left), modeled as a PS distribution $PS(\nu_\mathbf{c}, \kappa_\mathbf{c})$; An example of the grasp configuration (bottom right), where red and green forks represent collision and collision-free grasps, respectively, at a contact point; The graphical model (top right) illustrating the conditional dependency between variables.}
    \label{fig:diverse_grasp}
\end{figure}

\noindent \textbf{Grasp Configuration:} An output representation of our network, representing a set of grasps with different approach directions at a contact point $\mathbf{c}$ (Fig. \ref{fig:diverse_grasp}). 
We denote it as $C$ and formulate it as 
\begin{equation}\label{eq:grasp_configuration}
        C = (\mathbf{c}, \mathbf{b}, \{ \mathbf{a}_i \}_{i=1}^{n_r}, 
        \{\sigma_i\}_{i=1}^{n_r}, w, q).
\end{equation}
The approach vectors $\{ \mathbf{a}_i \}_{i=1}^{n_r}$ are derived from the $\mathbf{b}$. Each $\mathbf{a_i}$ is obtained by rotating an orthogonal vector $\mathbf{b_\perp}$ around $\mathbf{b}$ by a rotation angles ${\gamma_i}$. The vector $\mathbf{b\perp}$ is obtained using the Gram-Schmidt process. The angles ${\gamma_i}_{i=1}^{n_r}$ are calculated by dividing an angle range into $n_r$ equal steps. This formulation allows each approach vector to be expressed as a linear transformation of $\mathbf{b}$, as shown in the equation:
\begin{equation}\label{eq:approach_vector}
        \mathbf{a}_i = \mathbf{R}_{\gamma_i} \mathbf{b}_\perp, 
        \mathbf{b}_\perp = \mathbf{R}_\perp \mathbf{b},
\end{equation}
where $\mathbf{R}_{\gamma_i} $ is the rotation matrix of $\gamma_i$ and $\mathbf{R}_\perp$ is a constant rotation matrix. 

Each approach vector $\mathbf{a}_i$ is labeled by a binary collision-score $\sigma_i$, which indicates if the grasp $(\mathbf{c}, \mathbf{b}, \mathbf{a}_i, w)$ is collision-free. 
The antipodal quality $q$ measures the \textit{antipodality} (Eq. \ref{eq: ap_score}) of the contact pair 
$(\mathbf{c}, \mathbf{c} + \mathbf{b} w)$. 

\subsection{Supervised Learning}
The goal is to train a network, which takes a 3D normal grid $\mathbf{V}$ as input and infers a set of grasp configurations.
We use each grasp configuration together with an uncertainty $\kappa$ to parameterize the grasp distribution for each scene. Given the true grasp samples, the network is trained to predict optimal parameters that maximize the likelihood (Figure \ref{fig:detection_model}). 
We decompose the learning problem into learning the following two functions: 
\begin{equation}
        f_1: \mathbf{V}_i \to p_\mathbf{c} \qquad
        f_2: \mathbf{c} \to (\mathbf{b}, \kappa, \{\sigma_i\}_{i=1}^{n_r}, w, q) 
\end{equation}
The first function $f_1$ approximates the contact point distribution, which maps each voxel $\mathbf{V}_i$ of the 
input grid to the probability that the voxel contains a contact point. The second function $f_2$ then infers the antipodal quality $q$ and the grasp parameters given the contact points sampled from $f_1$. 
\section{METHODS}
\subsection{Grasp Distribution Modeling} 
We introduce a parameterized grasp distribution model to consider multiple grasp orientations and their orientation uncertainty at a contact point by representing the grasp distribution as a joint distribution over the grasp parameters $\mathbf{b}, \mathbf{a}, \mathbf{c}, \mathbf{t}, w$ conditioned on the input normal grid $\mathbf{V}$: 
\begin{equation}\label{eq:factorized_joint_dist}
    P(\mathbf{b}, \mathbf{a}, \mathbf{c}, \mathbf{t}, w | \mathbf{V}).
\end{equation}
Here, we incorporate the approach point locations denoted as $\mathbf{t}$ into our model, as they indicate the gripper hand's position relative to an approach direction, which is critical for assessing whether a grasp may lead to collisions. The grasp distribution can then be factorized into a contact and approach point distribution $P(\mathbf{c} | \mathbf{V})$ and the local grasp distribution $P(\mathbf{b}, \mathbf{a}, \mathbf{t}, w | \mathbf{c, V})$.

\subsubsection{Contact Point Distribution}
\noindent We represent each bin picking scene as a 3D label grid $L$ where each voxel belongs to one of the three non-overlapping sub-spaces: (1)~contact space of the grasp contact point locations; (2)~gripper space of the collision-free gripper poses represented by the approach point locations; (3)~empty space which is the complement of the contact space and the gripper space. 
Accordingly, each voxel is classified as contact voxel $l_\mathbf{c}$, approach point voxels $l_\mathbf{t}$, or empty voxel $l_\mathbf{e}$.
The contact point distribution is then approximated by the prediction of contact voxels.

\subsubsection{Local Grasp Distribution}
We further factorize the local grasp distribution into three distributions based on the conditional dependency between the grasp parameters as shown in Fig.~\ref{fig:diverse_grasp}. The factorization is formulated as 
\vspace{-0.1cm}
\begin{equation}
\begin{aligned}
    & P(\mathbf{b}, \mathbf{a},  \mathbf{t}, w|\mathbf{c})  = \int_\gamma P(\mathbf{b}, \mathbf{a},  \mathbf{t}, w, \gamma | \mathbf{c}) d \gamma \\
    & = P(w | \mathbf{b}, \mathbf{c}) P(\mathbf{b} | \mathbf{c}) 
    \int_\gamma P(\mathbf{a},  \mathbf{t}, \gamma | \mathbf{b}, w, \mathbf{c}) d \gamma \\ 
    &= 
    \resizebox{0.9\columnwidth}{!}{$
    \underbrace{P(w | \mathbf{b}, \mathbf{c})}_{\textrm{opening width}}
    \underbrace{P(\mathbf{b} | \mathbf{c})}_{\textrm{baseline}}
    \underbrace{\int_\gamma P(\mathbf{a} | \mathbf{b}, \gamma) 
    P(\mathbf{t} | \mathbf{b}, \mathbf{c}, \gamma) 
    P(\gamma | \mathbf{b}, \mathbf{c})d \gamma}_{\textrm{approach}}$}
    \end{aligned}
    \label{eq:factorized_grasp_dist}
\end{equation}
Here, we removed $\mathbf{V}$ just for a simplification.

\subsubsection{Baseline Vector Distribution} 
For a given contact point, there are multiple baseline vectors that represent a feasible grasp as shown in Fig.~\ref{fig:diverse_grasp}.
We model the distribution of baseline vectors at a contact point using the the Power-Spherical (PS) distribution \cite{de2020power} parameterized by the optimal baseline vector $\nu_\mathbf{c}$ and the concentration parameter $\kappa_\mathbf{c}$, formulated as 
\begin{equation}
    P(\mathbf{b} | \mathbf{c})=P_{PS}(\mathbf{b} | \nu_\mathbf{c}, \kappa_\mathbf{c}),
\end{equation}
where $P_{PS}(\cdot | \nu_\mathbf{c}, \kappa_\mathbf{c})$ is the probabilistic density function (PDF) of the PS distribution $PS(\nu_\mathbf{c}, \kappa_\mathbf{c})$.
The optimal baseline vector refers to the one that represents the most stable contact point pair with the highest antipodal grasp quality. 

\subsubsection{Approach Vector Distribution} 
Since the PS distribution is closed under linear transformation \cite{de2020power}, based on Eq.~\ref{eq:approach_vector}, approach vectors with respect to a rotation $\gamma$ follows the PS distribution $PS(\mathbf{R}_\gamma \mathbf{R}_\perp \mathbf{\nu_c}, \kappa_\mathbf{c})$.
Thus, we model the approach vector distribution as a mixture PS distribution based on Eq.~\ref{eq:factorized_grasp_dist} by defining the component and weight functions as 
\begin{equation}
    P(\mathbf{a} | \mathbf{b}, \gamma) := P_{PS}(\mathbf{a} |\mathbf{R}_\gamma \mathbf{R}_\perp \mathbf{\nu_c}, \kappa_\mathbf{c}),
\end{equation}
\begin{equation}\label{eq:weight_func}
    P(\mathbf{t} | \mathbf{b}, \mathbf{c}, \gamma, \mathbf{V})P(\gamma | \mathbf{b} , \mathbf{c}) := \sigma (\gamma).
\end{equation}
We assume that the weight function $\sigma(\gamma)$ is a binary piece-wise constant function, indicating whether an approach direction $\mathbf{a}_\gamma$ results in collisions.
Computing the continuous integral of this mixture model during our training is intractable. Hence, we approximate the function using $n_r$ binary values, corresponding to collision-scores $\{\sigma_i\}_{i=1}^{n_r}$ of $n_r$ rotation angles (Eq.~\ref{eq:grasp_configuration}).
We can then formulate the approach vector distribution at a contact point $\mathbf{c}$ as 
\begin{equation}
    P_\mathbf{c}(\mathbf{a}) = \frac{1}{\sum_{i=1}^{n_r} \sigma_i}\sum_{i=1}^{n_r} \sigma_i
    P_{PS}(\mathbf{a} | \mathbf{R}_{\gamma_i} \mathbf{R}_\perp \mathbf{\nu_c}, \kappa_\mathbf{c}).
\end{equation}

\subsection{Two-stage Network}
\begin{figure}[!t]
    \centering
    \vspace{0.3cm}
    \includegraphics[width= 0.95\linewidth]{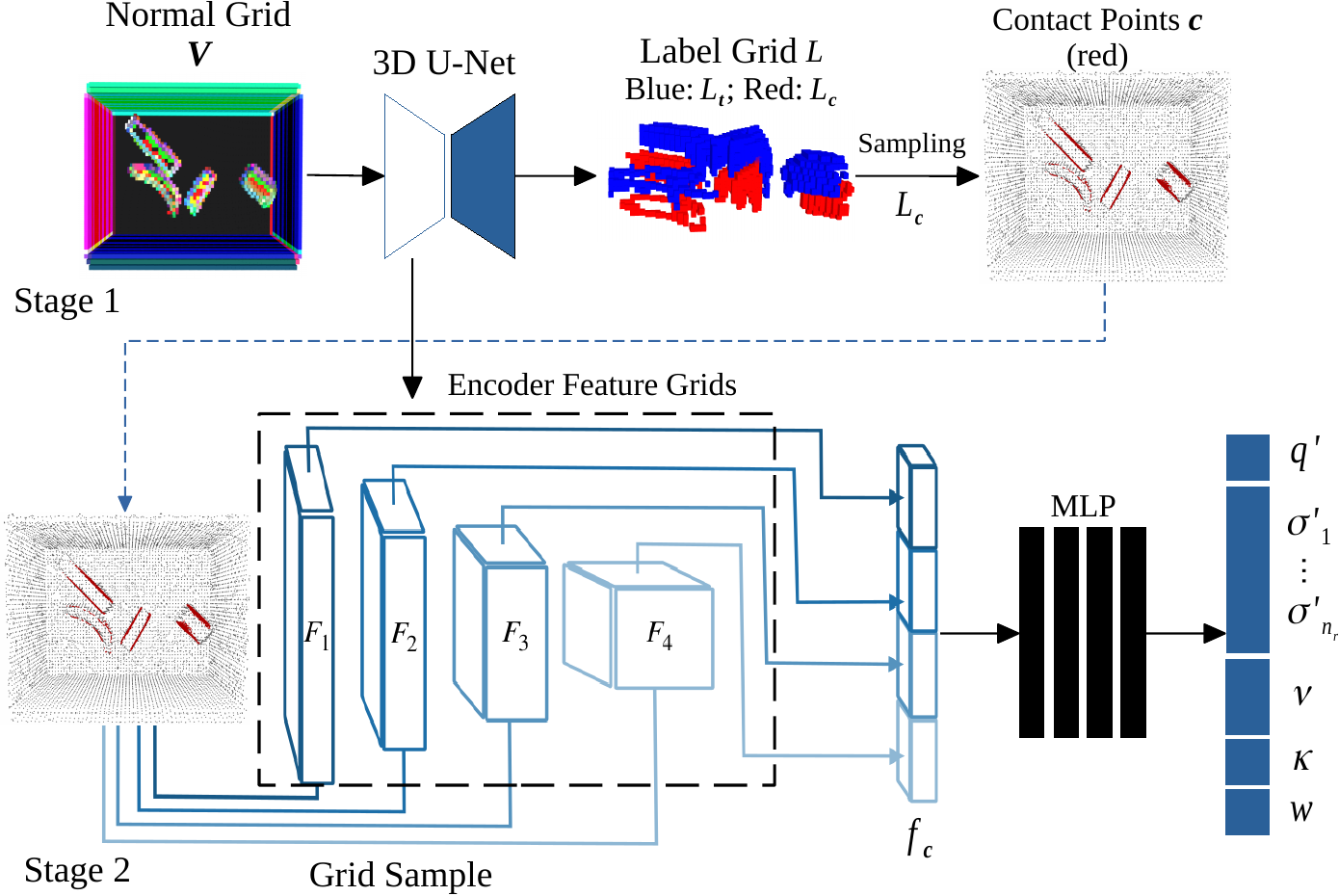}
    \caption{The two-stage network architecture. }
    \label{fig:network}
\end{figure}

Based on the grasp distribution modeling, we propose a two-stage neural network shown in Fig.~\ref{fig:network}. 

In the first stage, a 3D Residual U-Net~\cite{resunet,3dunet_pytorch} infers the label grid $\mathbf{L}$ given a 3D normal grid. Then, we sample $n$ contact points from contact voxels.
Each voxel of $\mathbf{L}$ stores the three confidence scores $[p_\mathbf{c}, p_\mathbf{t}, p_\mathbf{e}]$ for the classes of contact voxel $l_\mathbf{c}$, approach voxel $l_\mathbf{t}$ and empty voxel $l_\mathbf{e}$. We denote the grid storing $p_\mathbf{c}$ as $L_\mathbf{c}$ and the grid storing $p_\mathbf{t}$ as $L_\mathbf{t}$.   
 
In the second stage, according to $n$ contact point locations, we first grid sample and concatenate encoder features from each features grid $F_i$, which are intermediate outputs of U-Net encoder layers, forming $n$ multi-resolution encoder feature vectors $\mathbf{f}_\mathbf{c}$.  
Each feature vector is then passed into a Multi-Layer Perceptron (MLP) generating the optimal baseline vector $\nu_\mathbf{c}$, inverse concentration $\kappa'_\mathbf{c}$, opening width $w_\mathbf{c}$, as well as the prior antipodal quality $q'_\mathbf{c}$ and $n_r$ prior collision-scores $\sigma'_{1, \mathbf{c}}, \dots, \sigma'_{n_r, \mathbf{c}}$ for each contact point. 
We refer them as "prior" quality/score because they are conditioned on the predicted label grid. We compute the quality $q_\mathbf{c}$ by multiplying $q'_\mathbf{c}$ with the confidence score of the respective contact point $\mathbf{c}$, formulated as 
\begin{equation}
    q_\mathbf{c} = I(L_\mathbf{c}, \mathbf{c}) q'_\mathbf{c},
\end{equation}
where $I$ denotes the trilinear interpolation function. 
Similarly, a collision-score is calculated by multiplying its prior approach direction to the confidence score of the predicted approach points $\mathbf{t}_{(i, \mathbf{c})}$, formulated as 
\begin{equation}
    \sigma_{(i, \mathbf{c})} = I(L_\mathbf{t}, \mathbf{t}_{(i, \mathbf{c})})) \sigma'_{(i, \mathbf{c})},
\end{equation}
which aligns with the weight function Eq. \ref{eq:weight_func}.
To obtain the concentration parameter $\kappa$, we derive it from $\kappa'$ using the equation:   
\begin{equation}
    \kappa = \min ( \max (\frac{\kappa_0}{\kappa' + \epsilon}, \kappa_0), 4\kappa_0)
\end{equation}
where $\kappa_0$ is a tunable scale hyperparameter, and $\epsilon$ is used to avoid division by zero. In our training, we set $\kappa_0 = 25$.
\subsubsection{Loss Functions}
The loss function $\mathcal{L}$ is a weighted sum of a \textit{label loss} $\mathcal{L}_{l}$ (the first stage) and a \textit{grasp loss} 
$\mathcal{L}_g$ (the second stage), formulated as 
\begin{equation}
    \mathcal{L} = \phi_l \mathcal{L}_l + \phi_g \mathcal{L}_g, 
\end{equation}
where $\mathcal{L}_{l}$ is the focal loss \cite{focal_loss} for the label grid classification and $\phi_l$ and $\phi_g$ are weights of two losses, respectively. 
The grasp loss is represented as 
\begin{equation}\label{eq:grasp_loss}
    \mathcal{L}_g = \eta_b \mathcal{L}_b + \eta_a \mathcal{L}_a + \eta_w \mathcal{L}_w + 
    \eta_q \mathcal{L}_q, 
\end{equation}
where $\mathcal{L}_{b}, \mathcal{L}_{a}, \mathcal{L}_{w}, \mathcal{L}_{q}$ are losses 
from the predictions of baseline vectors, approach vectors, opening widths and grasp qualities at the sampled contact points, respectively and $\eta_b, \eta_a, \eta_w, \eta_q$ are corresponding weights. 

Baseline vector and approach vector loss functions are their negative log-likelihood functions.
The opening width loss $\mathcal{L}_w$ is minimum distance between predicted contact point pair to its true contact point pair. For a better training convergence, we use the estimated baseline vector $\hat{\nu}_\mathbf{c}$ to compute predicted approach vectors and contact pairs, which is an average over baseline vectors of neighbour grasp labels.
Moreover, $\mathcal{L}_{q}$ is the $L_1$ loss function. 

\subsection{Training \& Inference}
We employ a $64^3$ normal grid input with a voxel size of approximately $\SI{9}{\milli\meter}$. Our neural network is trained with a batch size of 1, a learning rate initialized at $10^{-3}$ and decayed exponentially to $10^{-5}$ at the training's end. The MLP in the second stage starts training after $100$ iterations of U-Net training. During training, we randomly sample a contact point within each voxel designated as a contact region. For inference, each contact region voxel is evenly subdivided into 8 sub-voxels. For each contact point, We generate 18 approach vectors. We define the rotation angle range as $[\frac{\pi}{2}, \frac{3\pi}{2}]$ to ensure that all approach vectors point downward.
To compute the loss, we assign weights of $10$, $5$, and $0.1$ to the label grid classes (contact space, gripper space, and empty space) for computing the label loss. In the grasp loss, we set the weights for $\mathcal{L}_{b}$, $\mathcal{L}_{a}$, $\mathcal{L}_{w}$, and $\mathcal{L}_{q}$ as $1$, $0.01$, $0.1$, $10$, and $0.1$, respectively. During training, we label each predicted contact point by searching for a maximum of 16 true contact points within a $\SI{3}{\milli\meter}$ radius. We compute $\mathcal{L}_{b}$, $\mathcal{L}_{a}$, and $\mathcal{L}_{w}$ only for contact points that have at least one true neighbor grasp sample.


\subsection{Data Generation} \label{sec:data_generation}
We create a synthetic 6-DoF grasp dataset to train our model, which contains diverse bin picking scenes along with corresponding dense grasp labels. We generate a total of $1000$ training scenes, each accommodating $1$ to $20$ objects. On average, each graspable object within these scenes is annotated with $470$ grasps. The simulator used for dataset generation is PyBullet \cite{coumans2021}.
Scenes are simulated by randomly dropping objects into the bin in a simulation environment. We select $200$ objects from the YCB \cite{calli2015ycb}, 
Google Scanned Objects (GSO) \cite{downs2022google} and ABC \cite{koch2019abc} datasets, splitting them randomly into training (149 objects) and testing (51 objects) sets. Objects are also randomly scaled during scene generation. Two different bin sizes are used for dataset creation. For each scene we render a top-down view depth images with a resolution of 640x480. We add noise to the rendered images in simulation using the same additive noise model applied in \cite{jiang2021synergies}. Afterwards, we generate grasp labels for each scene following the pipeline proposed in \cite{eppner2021acronym}. The difference is that we kept all valid grasps that are collision-free and have a high antipodal quality (above $0.5$) for each scene.

\section{EXPERIMENTS}
We conducted simulated and real-robot experiments to evaluate the performance of our grasp prediction model. 
\subsection{Simulated Experiments}
Simulated experiments were executed on hardware comprising an NVIDIA Corporation GP107GLM and an Intel(R) Core(TM) i7-8850H CPU. A standard Franka parallel-jaw gripper was employed.
To generate diverse test scenarios, we followed the training scene generation pipeline. We created 300 test scenes, evenly distributed among $easy$, $medium$ and $challenging$ scenarios, accommodating up to 5, 15, and 35 objects, respectively. $Easy$ scenes featured objects with simple structures (box, ball), whereas $medium$ and $challenging$ scenarios contain all objects from the test set.
Additionally, we add noise to $challenging$ depth images, replicating the conditions of generating training images. 
The dimensions of the bin model in each test scene were $600 \times 400 \times \SI{280}{\milli\meter}^3$.
Performance was evaluated based on a PyBullet \cite{coumans2021} simulation using the following metrics averaged over $100$ scenes in each scenario:
\begin{itemize}
    \item {\it Success Rate (SR)}: The ratio of successful grasps.
    \item {\it Clearing Rate (CR)}: The ratio of objects removed.
\end{itemize}

In each round, a grasp is executed if its antipodal quality and collision-score are greater than 0.5, and no collision at its pre-grasp pose. We test grasp configurations in a order of high to low antipodal quality. For each grasp configuration, we test first 8 out of 18 approach directions according to their collision-scores until find an executable grasp. A round ends when the bin is empty, no grasp is detected, or three consecutive failed attempts occur. 



\subsubsection{Baselines}
We compare our approach with the three baseline methods: 
\begin{itemize}
    \item GPD~\cite{ten2017grasp}: a two-stage 6-DoF grasp detection algorithm that generates a large set of grasp candidates sampled based on an input point cloud and evaluates each of them using a neural function.
    \item VGN~\cite{breyer2020volumetric}: a 3D convolutional neural network takes a TSDF grid as input and outputs grasps parameters at each voxel centers.
    \item GIGA~\cite{jiang2021synergies}: a two-stage network that jointly detects \mbox{6-DoF} grasp poses and reconstructs the 3D scene given a single-view TSDF grid.
\end{itemize}
GIGA and VGN were retrained on our dataset. To this end, we label each voxel containing at least one grasp center with the grasp orientation having the highest antipodal quality. We use the same top-down view depth images to test these baseline methods.
We additionally compare the performance to two ablated versions of our methods: "wo $\kappa$ + 1app and "w $\kappa$ + 1app".
For the first one, we trained our network to learn the optimal baseline vector by maximizing cosine similarity between each predicted baseline vector and the average vector over its neighbor ground truth baseline vectors, {\it without} considering the uncertainty $\kappa$. 
The second one is trained {\it with} the uncertainty learning. 
However, during experiments, we only return the optimal approach direction for each contact point. 

\subsubsection{Results}
\begin{figure}%
    \centering
    \subfloat{{\includegraphics[width=0.32\columnwidth]{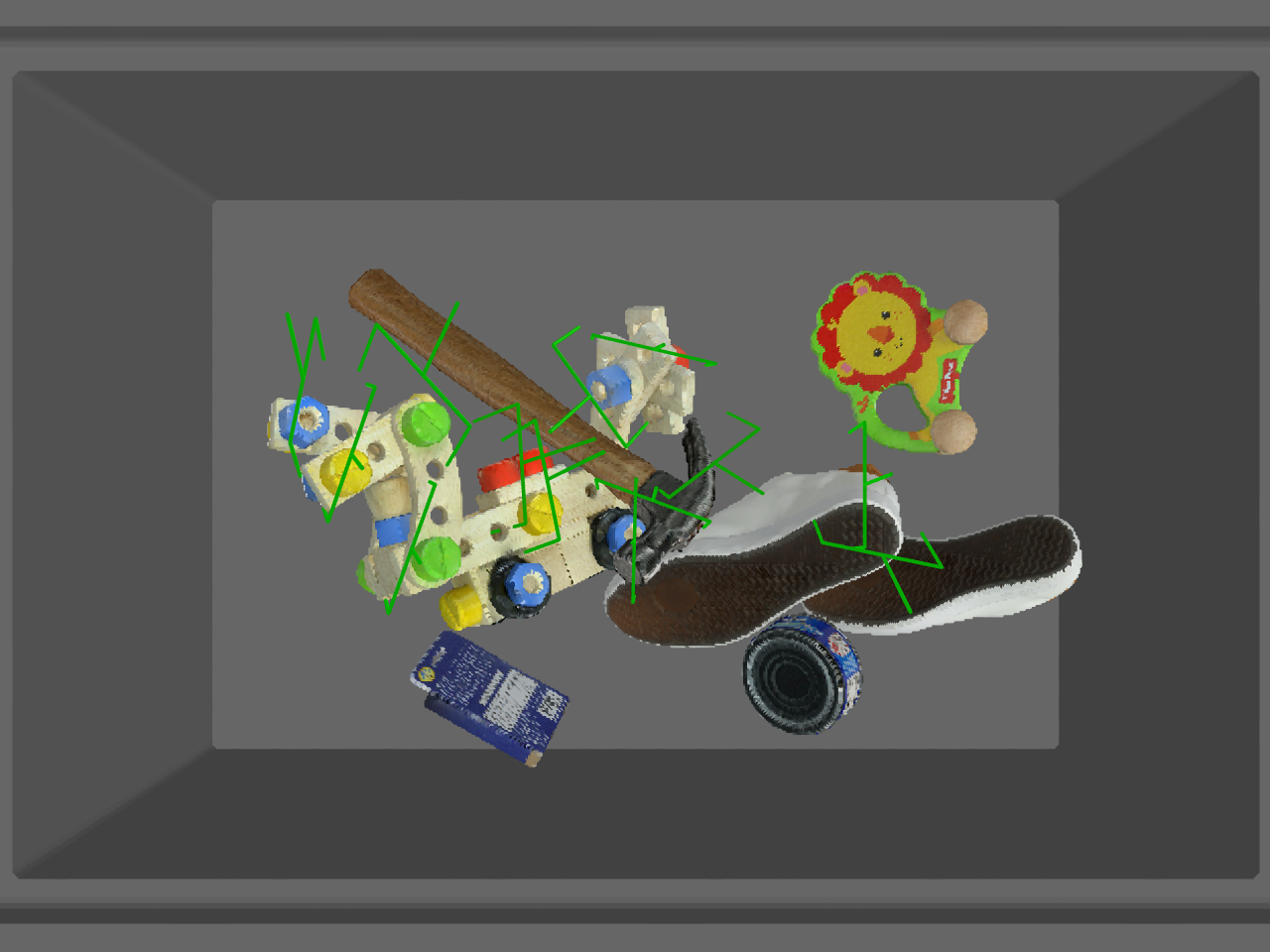} }}%
    \subfloat{{\includegraphics[width=0.32\columnwidth]{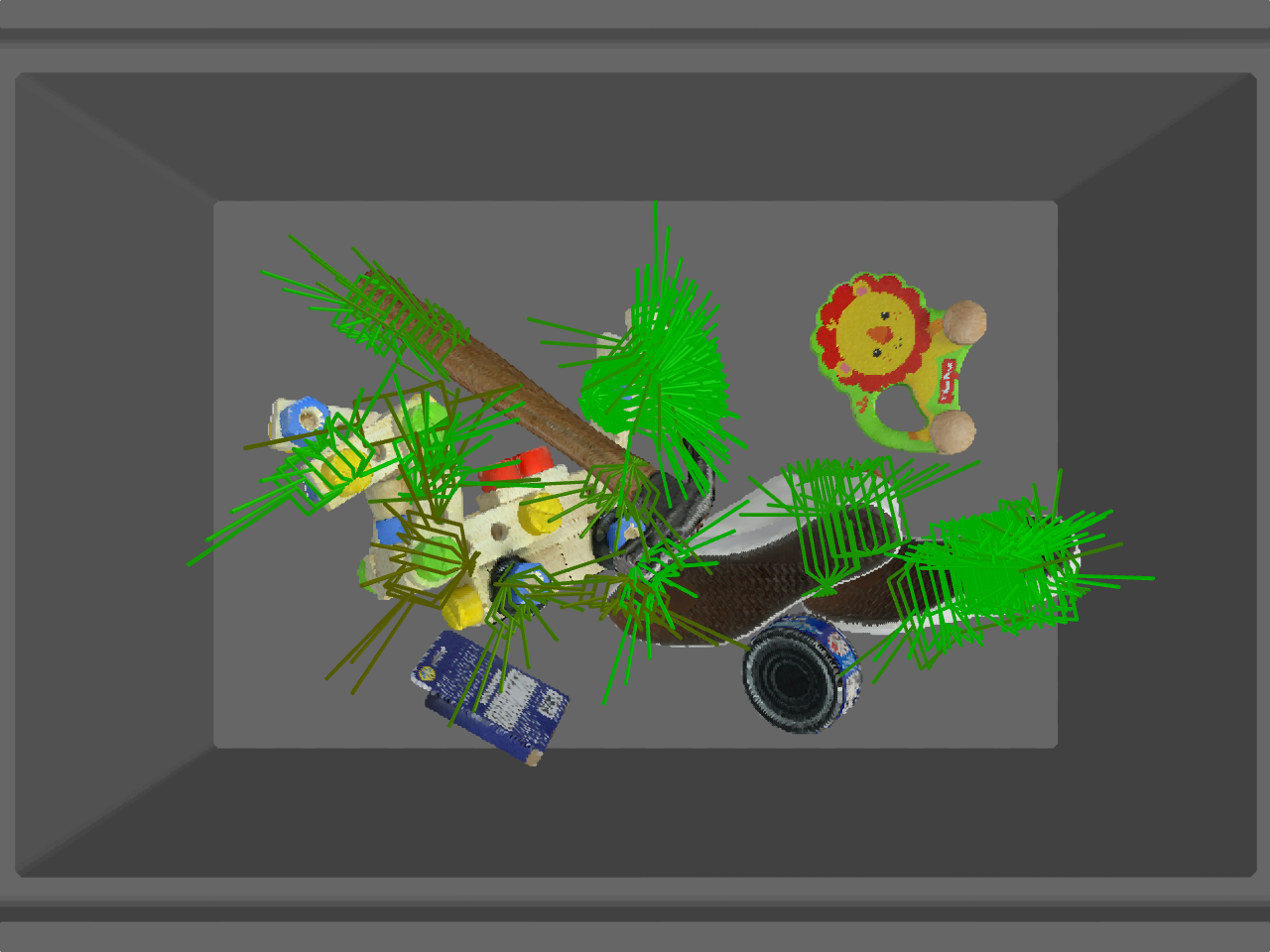} }}%
    \subfloat{{\includegraphics[width=0.32\columnwidth]{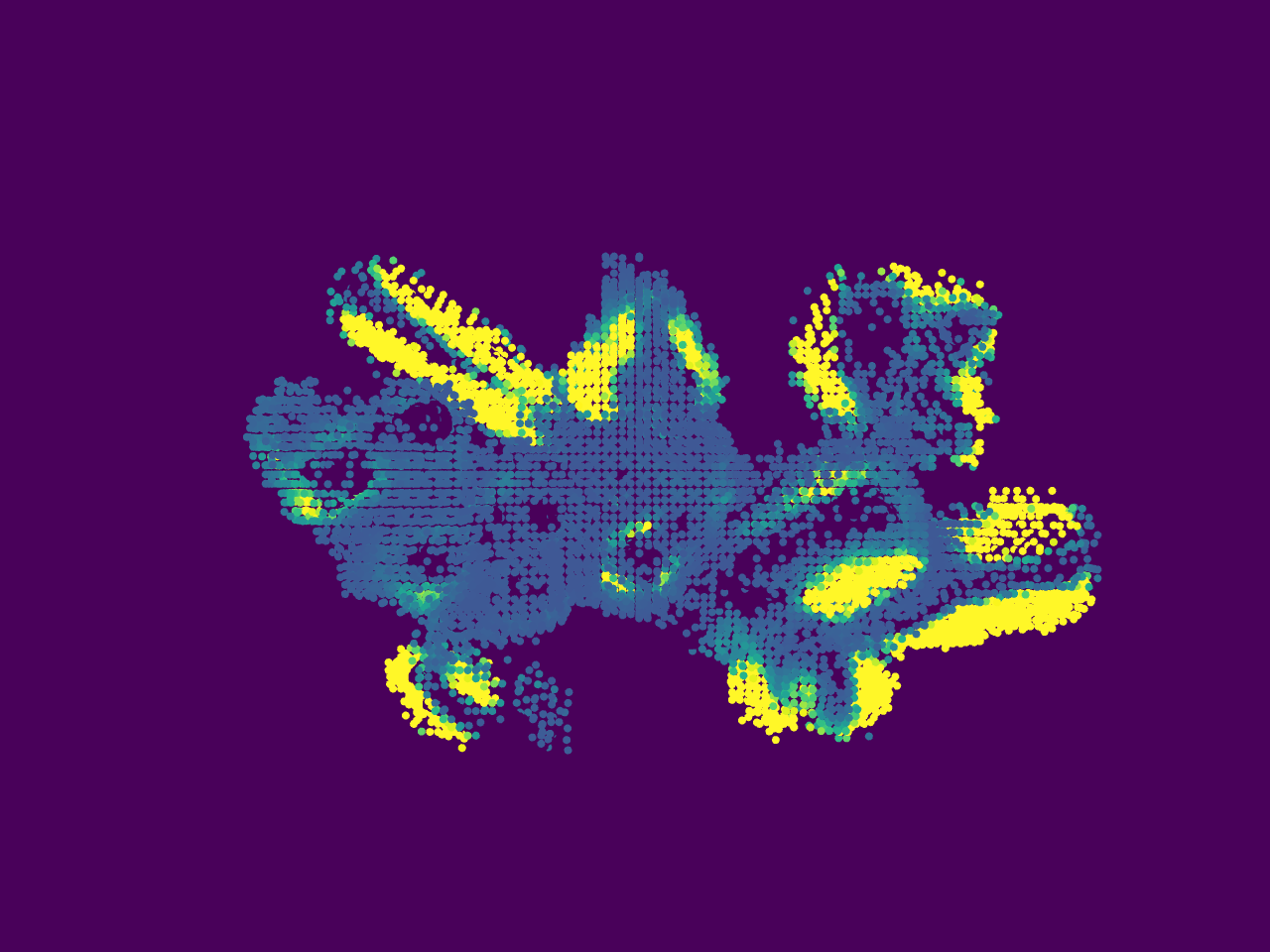} }}%
    \caption{Predicted grasps from GIGA (left), ours (middle), and our uncertainty prediction (right) on a $medium$ test scene.}
    \label{fig:results}%
\end{figure}

Our experimental results (Table I) across 100 test scenes in three scenarios demonstrate our method's robustness in managing cluttered and noisy input. Even in the most challenging scenario, it achieves $70\%$ success rate and $83\%$ clearing rate, outperforming the baseline methods. 
Even though "w $\kappa$ + 1app" only executes a single approach direction, it still generates more accurate and denser grasps than GIGA and VGN (see Fig.~\ref{fig:results}).
Additionally, we compare between models trained with and without uncertainty learning for the baseline vector prediction, specifically "w $\kappa$ + 1app" and "wo $\kappa$ + 1app". Our observations indicate a significant performance gap between the two, particularly in transitioning from easy to challenging scenarios. This suggests that uncertainty learning is crucial for improving grasp prediction accuracy and making the network more robust in noisy and cluttered input situations. In Fig.~\ref{fig:results}, we also present an example of the uncertainty prediction, which we can see that the model predict low uncertainty values (yellow points) to contact regions that are less cluttered and more easier to approach.
Lastly, we also observe the multi-approaches grasp prediction (8app) effectively handles collisions that occur when the gripper approaching the target object without prior motion planning. Importantly, this capability was not explicitly trained but emerges in practice. This is also reflected by the higher SR and CR when employing multi-approach grasps in medium and challenging scenarios, in comparison to using single-approach grasps.
In terms of runtime, our model has an inference time of approximately \SI{236}{ms} on our machine, while GIGA requires about \SI{233}{ms}, and VGN about \SI{136}{ms}. The runtime for GPD varies with difficulty levels, taking around \SI{373}{ms}, \SI{496}{ms}, and \SI{638}{ms} in three different scenarios.
\begin{table}[!h]
        \centering
        \resizebox{\columnwidth}{!}{%
        \def\arraystretch{1.2}\tabcolsep=5pt
        \begin{tabular}{@{}ccccccc@{}}
        \hline
        Method & \multicolumn{2}{c}{Easy}      & \multicolumn{2}{c}{Medium}    & \multicolumn{2}{c}{Challenging}      \\ \hline
               & SR (\%)      & CR (\%)       & SR (\%)      & CR (\%)       & SR (\%)      & CR (\%)       \\ \hline
        GPD \cite{ten2017grasp}   & 45.0          & 36.6          & 33.5          & 11.1          & 20.7          & 4.1           \\
        VGN \cite{breyer2020volumetric}   & 58.4          & 51.9          & 48.4          & 27.0          & 39.8          & 12.0           \\
        GIGA \cite{jiang2021synergies}  &  64.1             &    59.1           &          57.8     &   37.9            &        54.7       &       32.6        \\ \hline
        Ours (wo $\kappa$ + 1app) & 71.0 & 93.6 & 59.5 & 73.5 & 49.9 & 43.8 \\ 
        Ours (w $\kappa$ + 1app)  & 81.0 & 96.6 &  66.0 & 82.5 & 62.8 & 72.3 \\ 
        Ours (w $\kappa$ + 8app)  & \textbf{87.4} & \textbf{97.8} & \textbf{75.8} & \textbf{91.5} & \textbf{70.5} & \textbf{83.0} \\ \hline
        \end{tabular}
        }
        \caption{Averaged success rate (SR) and clearing rate (CR) over 100 test scenes in $easy$, $medium$ and $challenging$ scenarios. Bold text indicates the optimal performance.} \label{tab:sim_results}
\end{table}

\subsection{Real Robot Experiments}
\subsubsection{Experiment Setup}
\begin{figure}%
    \centering
    \vspace{0.3cm}
    \includegraphics[height=4.0cm]{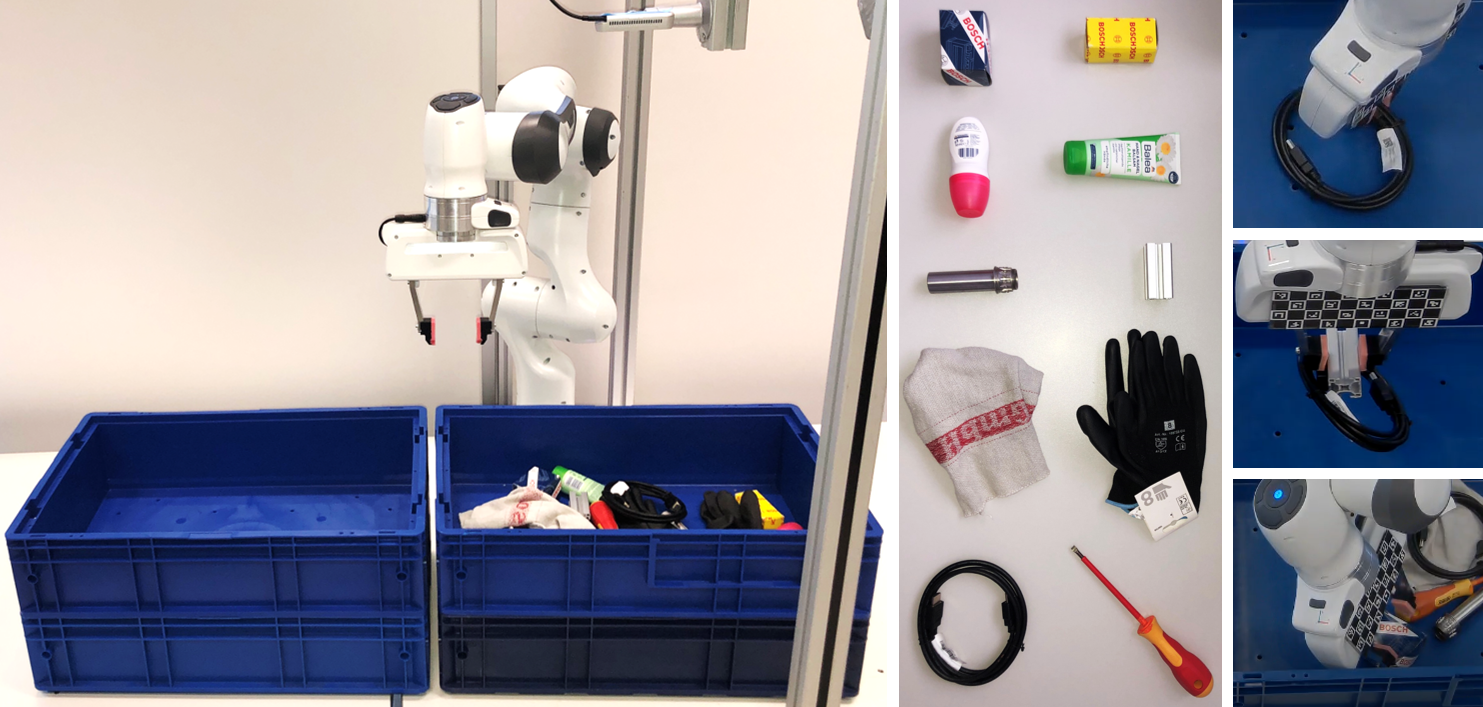}
    \caption{Experimental setup with robotic arm, parallel-jaw gripper, overhead RGBD camera and bins (left), diverse object portfolio (middle), successful grasp (top right), failure grasp (middle right), collision grasp (bottom right).} 
    \label{fig:real_setup_objects}%
\end{figure}

In the real robot experiments, we used a 7-axis Franka Emika Panda robot equipped a 
Franka parallel-jaw gripper with custom silicon fingers (see Fig.~\ref{fig:real_setup_objects}).
A RealSense D415 RGBD camera with a resolution of $1920 \times 1080$ is statically mounted above a source bin with an distance of around $\SI{0.6}{\meter}$ for top-down view image capturing of the bin picking scene. The test object portfolio consists of 10 diverse objects including boxes, tubes, industrial objects, textiles, a USB cable and a screw driver.
In each round, we place all test objects in the source bin an perform a manual mixing to generate random object poses within the bin. 
Additionally, we remove objects close to the bin walls that would not be graspable due to collisions between the used gripper and the bin and drop them manually from a random position above the bin.
We use the same model weights and network configuration as in the simulated experiments.
During the experiments, we determine the final grasp among all grasp predictions by selecting the grasp configuration with the highest grasp quality score. 
Then we select the grasp with the lowest collision score within the previously selected grasp configuration.
In case of collisions between gripper and bin or objects during a grasp attempt, we first manually stop the robot and then consider the next highest predicted collision-score for the selected grasp configuration. With this setting, we run 10 rounds in total. A round stops if we had 3 consecutive failure grasps. 

\subsubsection{Results}
The robot experiments result in a clearing rate of $90 \%$ and a success rate of $77 \%$ with a total number of $116$ grasp attempts and $90$ successful grasps. We also encountered 4 double grasps where two objects were grasped at once during a grasp attempt.
In this case, the grasp counts as a failed attempt and we dropped the objects in the source bin from a random position above the bin.
The results show a similar performance as in the simulated experiments without any adjustments of the trained model which indicates the robustness to noisy image input data due to the grasp distribution modeling.




\section{CONCLUSIONS}
    In this work, we considered the problem of 6-DoF grasp detection for cluttered and unknown objects located inside a bin.
    We presented a novel approach to predict dense and collision-free grasps with multiple orientations per grasp point based on two-stage network architecture.
    The approach includes a new method for grasp distribution modeling based on Power Spherical distributions that accounts for grasp orienation uncertainty and enables a training based on multiple orientations per grasp contact.
    Hence, the network can be trained with more potential ground truth grasps and allows to predict more feasible grasps which is essential when grapsing with limited reachability inside a bin.
    Using only a single-view depth image we can predict dense and collision-free grasps on cluttered object scenes including multiple grasp orientations at grasp point.
    We evaluated our approach with simulated and real robot experiments on diverse object types resulting in a clearing rate of around 90\%.
    We also showed that we outperform state of the art approaches for 6-DoF grasp prediction in bin picking tasks.
    Future work could consider diverse gripper designs in the dataset generation or the network architecture to enable the usage for different robot setups.
    Moreover, the approach might be combined with additional robot pushing skills or advanced grasp selection methods to further increase the object clearing performance in challenging bin picking scenarios.

\bibliographystyle{IEEEtran}
\bibliography{6dof_parallel_grasp}

\end{document}